\documentclass[10pt, a4paper]{article}
\usepackage[final]{lrec-coling2024} 
\usepackage{tabularray}
\usepackage{amsmath}
\usepackage{graphicx}
\usepackage{booktabs}
\usepackage{subcaption}
\usepackage{adjustbox}
\usepackage{pifont}
\usepackage{multirow}
\usepackage{microtype}
\usepackage{hhline}
\usepackage{setspace}
\usepackage{multirow}
\usepackage{booktabs}
\usepackage{amssymb}
\newcommand{\ngrams}{$n$-gram }

\title{\textbf{Qsnail: A Questionnaire Dataset
for Sequential Question Generation}}

\name{Yan Lei$^{1,2}$, Liang Pang$^{1}$\sthanks{\ \ Corresponding authors}, Yuanzhuo Wang$^{1}$, Huawei Shen$^{1}$, Xueqi Cheng$^{1}$}
\address{
    $^{1}$CAS Key Laboratory of AI Safety \& Security,\\
  Institute of Computing Technology, Chinese Academy of Sciences, Beijing, China, \\
    $^{2}$University of Chinese Academy of Sciences, Beijing, China \\
    \texttt{\{leiyan21b, pangliang, wangyuanzhuo, shenhuawei, cxq\}@ict.ac.cn}
    }

\abstract{
The questionnaire is a professional research methodology used for both qualitative and quantitative analysis of human opinions, preferences, attitudes, and behaviors. 
However, designing and evaluating questionnaires demands significant effort due to their intricate and complex structure.
Questionnaires entail a series of questions that must conform to intricate constraints involving the questions, options, and overall structure.
Specifically, the questions should be relevant and specific to the given research topic and intent.
The options should be tailored to the questions, ensuring they are mutually exclusive, completed, and ordered sensibly. 
Moreover, the sequence of questions should follow a logical order, grouping similar topics together. 
As a result, automatically generating questionnaires presents a significant challenge and this area has received limited attention primarily due to the scarcity of high-quality datasets. 
To address these issues, we present Qsnail, the first dataset specifically constructed for the questionnaire generation task, which comprises 13,168 human-written questionnaires gathered from online platforms.
We further conduct experiments on Qsnail, and the results reveal that retrieval models and traditional generative models do not fully align with the given research topic and intents. Large language models, while more closely related to the research topic and intents, exhibit significant limitations in terms of diversity and specificity. Despite enhancements through the chain-of-thought prompt and finetuning, questionnaires generated by language models still fall short of human-written questionnaires. Therefore, questionnaire generation is challenging and needs to be further explored. The dataset is available at: ~\url{https://github.com/LeiyanGithub/qsnail}. 
\\ \newline 
}

\begin{document}
\maketitleabstract

\section{Introduction} \label{introduction}
The questionnaire serves as a professional research tool designed for gathering data on human opinions, attitudes, preferences, and behaviors~\cite{dyer1976questionnaire, hammarberg2016qualitative}. Typically, when given a specific research topic and intents, a questionnaire is designed to gather information from respondents. The research topic and intents are the ideas about what kind of information wants to be collected. Surpassing the limitations of a single voting question, questionnaires consist of sequential questions, enabling comprehensive qualitative and quantitative analyses, thereby fostering more profound and convincing conclusions or suggestions~\cite{ponto2015understanding}. Consequently, they are extensively utilized in diverse domains such as education, healthcare, government, and psychology~\cite{artino2014developing}. 

Questionnaire design is a systematic process involving background research, question formulation, option configuration, sequence adjustment, pre-testing, and other multiple stages~\cite{ krosnick2018questionnaire}. This process demands significant domain knowledge and cognitive effort, requiring multiple human efforts and time. 
Pre-trained language models (PLMs) based on Transformers~\cite{vaswani2017attention} have achieved great success on text generation tasks, including creative writing, question answering~\cite{xu2023search, deng2023regavae}, and dialogue generation~\cite{zhou2023simoap, yang2023multi, ouyang2022training, chowdhery2022palm}. Especially, ChatGPT~\cite{ouyang2022training} can generate high-quality content following user instructions, which makes generating usable questionnaires possible. This leads us to question how well these models perform in generating questionnaires.

\begin{figure}
    \includegraphics[width=0.48\textwidth]{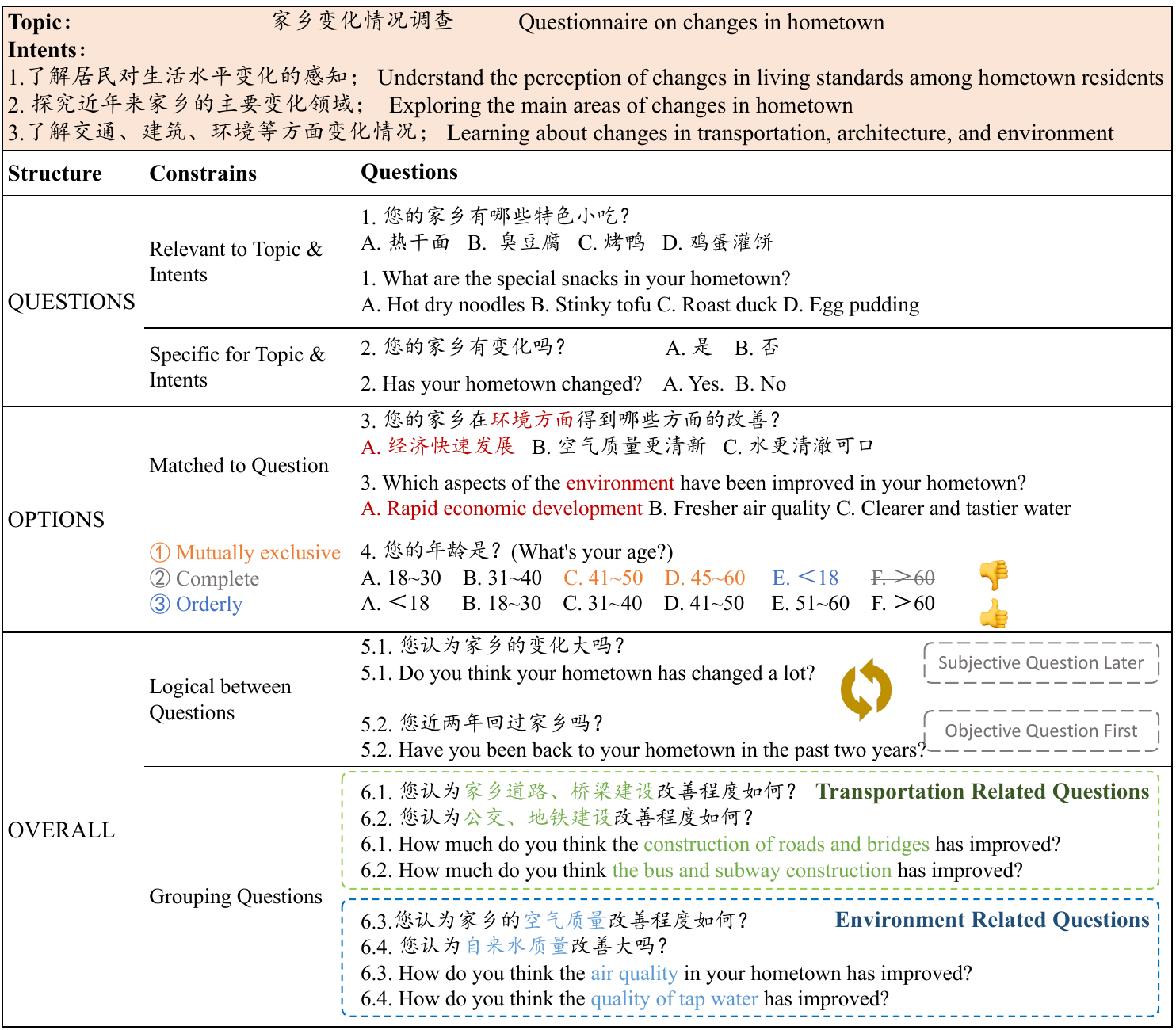}
    \caption{An example of intricate constraints in the questionnaire on the research topic: \textit{changes in hometown}. The top is about the research topic and intents of the questionnaire, while the left side is about questions, options, and overall constraints.}
    \label{fig:fig1}
\end{figure}

To date, unlike previous sequential question generation works that aim at resolving the coreference alignment, the sequential questions in the questionnaire focus much on inherent constraints that can be classified into questions, options, and overall aspects, as depicted in Figure~\ref{fig:fig1}. \textbf{The individual question} should be 
(1) \textit{Relevant to the Topic and Intents}: the questions must serve the research targets. The question ``Snacks in hometown'' is not related to the topic ``Changes in my hometown'', so it should be removed. 
(2) \textit{Specific for the Topic and Intents}~\cite{chiang2015constructing, lee2006constructing}: abstract questions are often vague and provide limited useful information. 
The question, ``Has your hometown changed?'', is ambiguous in terms of time scale and perspective, making it difficult to judge whether to answer ``yes'' or ``no''.
\textbf{The options of the question} must be 
(1) \textit{Matched with the question}: option A (rapid economic development) of question 3 in Figure~\ref{fig:fig1} does not match the environmental change aspect of the question. 
(2) \textit{Mutually exclusive, complete, and orderly}~\cite{taylor1998questionnaire}: for instance question 4 in Figure~\ref{fig:fig1}, 
the inclusion of age 46 in both options C (41--50) and D (45--60) violates the principle of exclusivity. Additionally, the absence of any option related to ages exceeding 60 violates the requirement for completeness. Furthermore, the presence of unsorted options, such as option E (<18), adds difficulty in identifying the appropriate choice.
\textbf{The order of sequential questions} should be 
(1) \textit{Logical}: from objective to subjective to keep a better logical flow. 
For instance, question 5.2 in Figure~\ref{fig:fig1} should be placed before question 5.1 since it requires individuals who have returned to their hometown to discuss any alterations that have taken place there.
(2) \textit{grouping similar topics together}~\cite{taherdoost2022designing}: for example in the last part in Figure~\ref{fig:fig1}, questions 6.1 and 6.2 both address changes in hometown transportation and should therefore be arranged together. Similarly, questions 6.3 and 6.4, which focus on environmental changes, should be grouped similarly. 
Considering the above constraints, automatic generation of questionnaires is a crucial yet challenging task. Until now, this area has received limited attention mainly due to the lack of high-quality datasets.

To emphasize the aforementioned challenges, we introduce the Qsnail dataset, a questionnaire collection established through a comprehensive process involving web crawling, data filtering, and intent reconstruction.
Specifically, we first collect the dataset by web crawling questionnaires from the Wenjuanxing and Tencent Wenjuan online form-filling platforms. Subsequently, we eliminate unqualified and duplicate data using keyword-based and md5 filtering mechanisms. We further ascertain the research intentions by leveraging the ChatGPT model to analyze the questionnaire content. As a result, Qsnail contains 13,168 high-quality human-written questionnaires, including approximately 184,854 question-option pairs and spanning 11 distinct application domains.

Moreover, it is worth noting that conventional text generation metrics are insufficient for evaluating this novel task. Consequently, we propose novel automatic and human evaluation metrics adapted to this task. Furthermore, we conduct comprehensive experiments to analyze the performance of both retrieval and generative models in this task. Our findings reveal that retrieval models often exhibit deviations from the research topic and intents, while traditional generative models face challenges in producing coherent and usable questionnaires, even after finetuning. Large language models, like ChatGPT, ChatGLM-6B, and Vicuna-7B, although highly relevant to the topic and intentions, display significant gaps in terms of diversity, specificity, rationality, order, and background when compared to human-level performance.
To alleviate the aforementioned shortcomings, we explore an outline-first prompt method and finetuning models, which yield improvements in specificity and rationality. Nevertheless, there remains a substantial disparity with humans in terms of diversity and specificity. Consequently, the task of questionnaire generation proves to be challenging and warrants further investigation.
Our contributions include:
\begin{itemize}
    \item Formalizing the questionnaire generation as a sequential question generation task and pointing out its challenges.
    \item Proposing a new questionnaire generation dataset, e.g. Qsnail, to involve more researchers focusing on this problem.
    \item To thoroughly assess the effectiveness of various models in this task, furthermore, explore two distinct approaches: the outline-first prompt and the model fine-tuning. 
    
\end{itemize}

\section{Qsnail Dataset}
The first phase entails setting a benchmark for the questionnaire generation task. Therefore, within this section, we provide detailed insights into the formulation of the questionnaire generation task, outline the process of dataset collection, and perform further analyses.

\subsection{Task Formulation}
Questionnaires consist of a series of interconnected questions designed to fulfill the specific research intents. The fundamental challenge in crafting questionnaires lies in the transformation of these research intents into a set of precise and targeted questions.
Typically, the input includes a research topic denoted as $T$ and a research intent represented as $I$. The research intent, in turn, encompasses various sub-targets, denoted as $i_1, i_2, \cdots, i_p$, described in natural language, where the value of $p$ corresponds to the number of sub-targets. A research topic is a particular concept or event the researcher wants to explore. The research intent outlines the precise information the researcher seeks to collect from respondents through the questionnaire. Notably, the design of questionnaires can vary significantly for the same research topic depending on the distinct research intents.
To illustrate, when examining a broad research topic like ``food preferences'', if the research objectives are general public, the aim is to gain insights into individuals' food preferences, their underlying motivations for liking or disliking specific foods, and relevant influencing factors. Conversely, when the research objective is centered on exploring the perspectives of chefs and food critics, the intent shifts to comprehending their unique motivations and criteria for assessing food.
Different research intents result in distinct questionnaires for the same topic. Consequently, a more detailed description of the research intents becomes crucial. Hence, the questionnaire generation task involves providing the research topic $T$ and intents $I$ as the inputs, which then generates a sequence of questions $Q_1, Q_2, ..., Q_m$, where $m$ denotes the total number of questions. Questions within the questionnaire can be divided into open-ended or closed-ended questions. $Q_i = \{q_i\}$ is open-ended question and $Q_i = \{q_i,o_1, o_2, \cdots, o_{n_i}\}$ is closed-ended question where additional options $o_j$ are attached, and $n_i$ denotes the number of options. Each individual question, along with its options, and the order of sequential questions must adhere to satisfy the constraints mentioned in Section~\ref{introduction}.

\subsection{Data Collection}
To ensure the acquisition of high-quality questionnaire data, our initial step involves crawling questionnaire data from two sources: Wenjuanxing~\footnote{\url{https://www.wjx.cn/}} and Tencent Wenjuan~\footnote{\url{https://wj.qq.com/}} platforms. However, due to the presence of numerous non-questionnaire forms and duplicated data, we implement a filtering mechanism utilizing keywords and MD5 hashing. Additionally, we employ the questionnaire contents to extract and reconstruct the underlying research intents. Details are provided below.

\textbf{Web Crawling}. Wenjuanxing and Tencent Wenjuan are extensively utilized for various purposes, including survey research, exam administration, and online voting. These tools serve to fulfill the data collection and statistical analysis requirements of diverse user groups, encompassing government agencies, educational institutions, and others.
Consequently, we construct a dataset containing human-written questionnaires spanning various research domains. Initially, we crawl approximately 30,000 random instances.

\textbf{Data Filtering}. The aforementioned platforms not only facilitate the creation of questionnaires but also provide other form-creation functions, including exam forms. However, this diversity in form types introduces noise into the collected data. To address this issue, we implement a two-step post-processing approach to extract clean data.

    (1) \textit{Keywords Filtering}. These platforms contain a considerable amount of non-survey forms, making it challenging to isolate pure questionnaires. Typically, questionnaires are distinguished by titles containing keywords such as         ``questionnaire'', ``survey'', or ``investigation'', Therefore, to ensure standardized questionnaire data, we apply strict keyword filtering. Furthermore, we remove any data containing personal private information during the filtering process. It is also important to note that the data is publicly available so they already undergo privacy and security audits by the platforms. 

    (2) \textit{MD5-based Filtering}. We also notice a significant presence of directly copied duplicate questionnaires. To preserve data diversity, we implement an MD5-based filter to eliminate duplicates.

Ultimately, we acquire 13,168 questionnaires, comprising a total of 184,854 question-option pairs. Each questionnaire comprises a title, a series of sequential questions, and their corresponding options. To guarantee data quality and privacy safety, we conduct a random sampling of 100 data examples for manual review, identifying only 5 instances that do not meet our criteria. 


\textbf{Intent Reconstruction}. The data retrieved from the online questionnaire website includes only the title and the corresponding questionnaire content. The creation of a specific questionnaire related to the research topic is unfeasible in the absence of explicit research intents. 
A description of the research intents is crucial. Typically, the research intent comes from the designer's initial thoughts before crafting it. 
Unfortunately, reaching out to the designers for clarification is often not possible. Another aspect that can shed light on the research intent is the sequence of questions and options in the questionnaire. 
Thus, we can conclude the research intent from the questionnaire content. 
However, manual annotation of this information is labor-intensive and time-consuming. Recent studies have utilized ChatGPT for various labeling tasks~\cite{gilardi2023chatgpt, tornberg2023chatgpt}. Consequently, we employ ChatGPT to generate research intents based on specific question-option pairs, limiting the output to no more than five sub-targets for practicality.
To ensure the reliability of these generated research intents, we randomly select 50 cases and engage three graduate students to perform manual evaluations. Our evaluation criteria consist of three key aspects: relevance, recall, and abstraction.
Relevance ensures that the model-generated research intents align with the original question-options pairs while avoiding unmentioned intents. Recall aims to encompass as many questions from the original questionnaire as possible, and abstraction requires that the research intent can condense multiple questions.
Relevance, recall, and abstraction are scored on a scale of 1 to 5, where 1 represents extremely poor, and 5 indicates extremely good. Human evaluation of the research intents scores 4.94 for relevance, 4.32 for recall, and 4.36 for abstraction. This demonstrates that the model-generated research intents closely align with our expectations in terms of relevance. However, they achieve acceptable levels of recall and abstraction.
It is important to highlight that the recall and abstraction metrics can be influenced by the number of questions within the questionnaire. With an abundance of questions, even for humans, condensing all survey intents into a limited number of sentences becomes challenging. Conversely, when the questionnaire contains only a few questions, each question aligns with a specific survey intent, potentially affecting abstraction metrics adversely.

\subsection{Data Analysis} 

\begin{table}
\caption{Comparison of Qsnail with other selected question generation datasets.}
\label{tab: Qsnail}
\adjustbox{max width=0.48\textwidth}{\begin{tabular}{@{}lccccc@{}}
\toprule
\multicolumn{1}{c}{} & Non-factoid & Logical & Grouping & Options & Scale \\ \midrule
SQuAD  ~\cite{DBLP:conf/acl/RajpurkarJL18}                & \ding{55}           & \ding{55}       & \ding{55}        & \ding{55}       & 230K  \\
LearningQ  ~\cite{chen2018learningq}            & \ding{55}           & \ding{51}       & \ding{55}        & \ding{55}       & 230K  \\
CoQA  ~\cite{DBLP:journals/tacl/ReddyCM19}                 & \ding{55}           & \ding{51}       & \ding{55}        & \ding{55}       & 127K  \\
QuAC  ~\cite{DBLP:conf/emnlp/ChoiHIYYCLZ18}                 & \ding{51}           & \ding{51}       & \ding{55}        & \ding{55}       & 100K  \\
RACE  ~\cite{DBLP:conf/emnlp/LaiXLYH17}                 & \ding{55}           & \ding{55}       & \ding{55}        & \ding{51}       & 97K   \\
MCTest  ~\cite{DBLP:conf/emnlp/RichardsonBR13}                  & \ding{55}           & \ding{55}       & \ding{55}        & \ding{51}       & 7K    \\
Qsnail                & \ding{51}           & \ding{51}       & \ding{51}        & \ding{51}       & 13K   \\ \bottomrule
\end{tabular}}
\end{table}

\textbf{Comparisons with Existing Datasets}. 
Most of the existing datasets used for question generation, as outlined in Table~\ref{tab: Qsnail}, primarily emphasize the creation of fact-based questions. For example, the questions within the SQuAD dataset are constructed by crowdsourced individuals with the primary aim of extracting factual information from source documents. Nevertheless, these datasets are not well-suited for the specific task of creating questionnaires. This unsuitability arises from two fundamental factors. Firstly, questionnaires consist of subjective inquiries, as opposed to the objective nature of the questions in these datasets. Secondly, questions in questionnaires feature a complex structure with intricate constraints.
The complexity of questionnaires becomes particularly evident in the case of sequential questions, each of which may be accompanied by numerous response options. Furthermore, these intricate constraints encompass various aspects, including questions, options, and overall levels. Consequently, Qsnail holds substantial promise for advancing research in the domain of automatic question generation.


\begin{table}
\caption{Statistics on Qsnail dataset. Count: the number of questionnaires; Len: average count of words per questionnaire/question/choice; $\overline{\textrm{Num}}$: average number per questionnaire/question;}
\label{tab: statistic}
\adjustbox{max width=0.48\textwidth}{\begin{tabular}{@{}ccccccc@{}}
\toprule
\multicolumn{2}{c}{Questionnaire} & Title & \multicolumn{2}{c}{Question} & \multicolumn{2}{c}{Choice} \\ 
Count             & Len             & Len   & $\overline{\textrm{Num}}$        & Len       & $\overline{\textrm{Num}}$          & Len         \\ \midrule
13168             & 596.5             & 15.8   & 14.2        & 20.4       & 4.4           & 5.1         \\ \bottomrule
\end{tabular}}
\end{table}

\begin{figure}
  \centering
  \begin{minipage}[b]{0.23\textwidth}
    \centering
    \includegraphics[width=\textwidth]{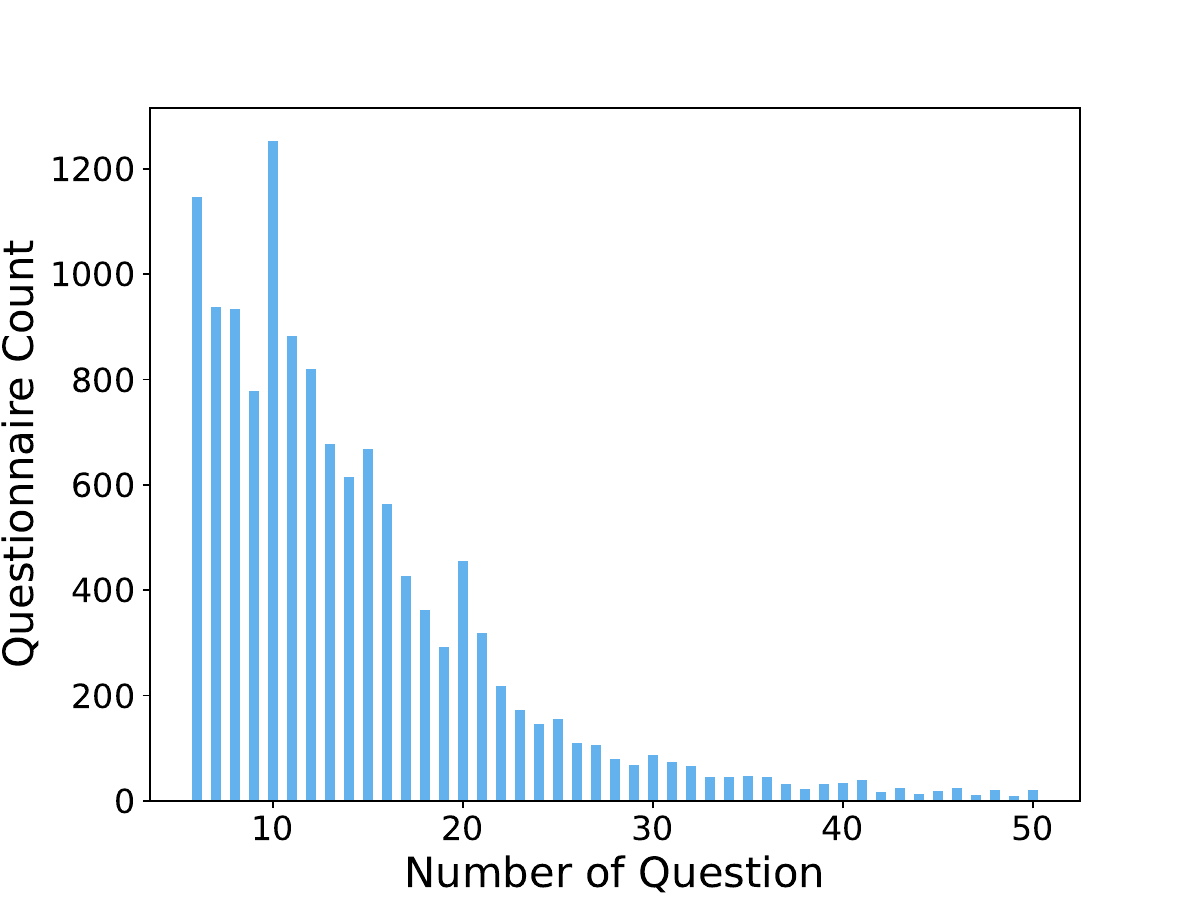}
    \subcaption{\scriptsize Question Number Statistic}
    \label{fig: question_num}
  \end{minipage}
  \begin{minipage}[b]{0.23\textwidth}
    \centering
    \includegraphics[width=\textwidth]{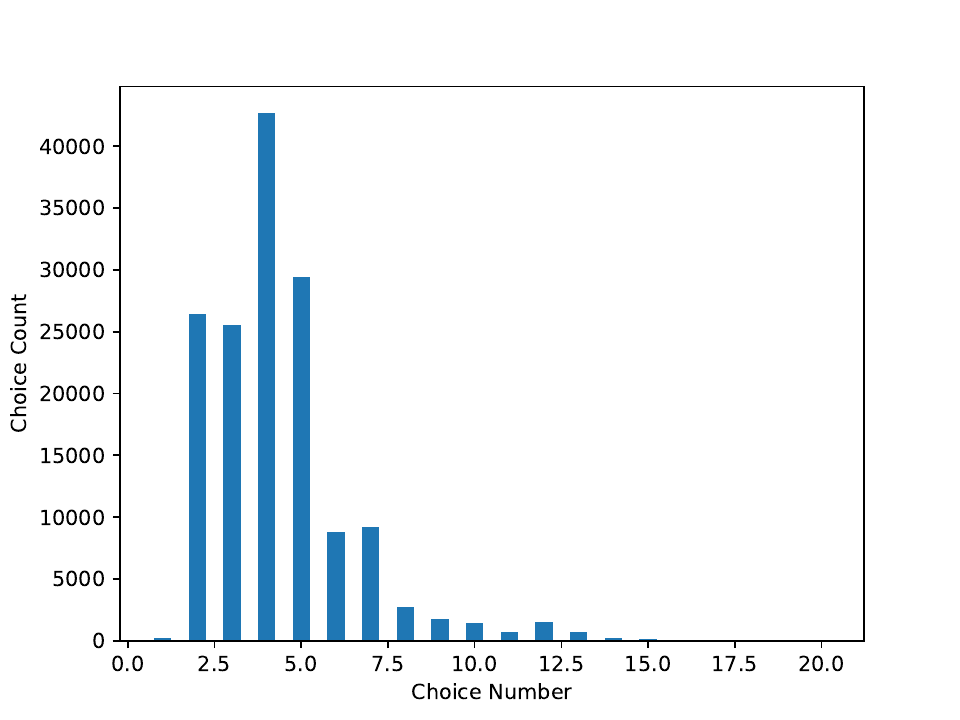}
    \subcaption{\scriptsize{Choice Number Statistic}}
    \label{fig: choice}
  \end{minipage}
  \begin{minipage}[b]{0.23\textwidth}
    \includegraphics[width=\textwidth]{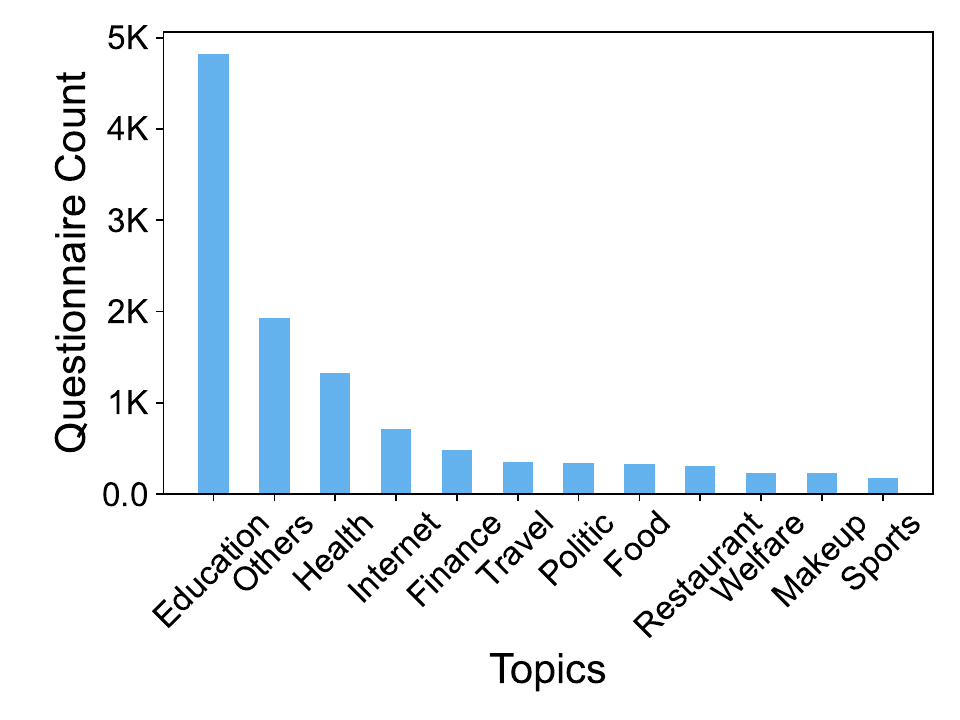}
    \subcaption{\scriptsize{Topic Distribution}}
    \label{fig: topic category}
  \end{minipage}
  \begin{minipage}[b]{0.23\textwidth}
    \centering
    \raisebox{0.10\height}{\includegraphics[width=\textwidth]{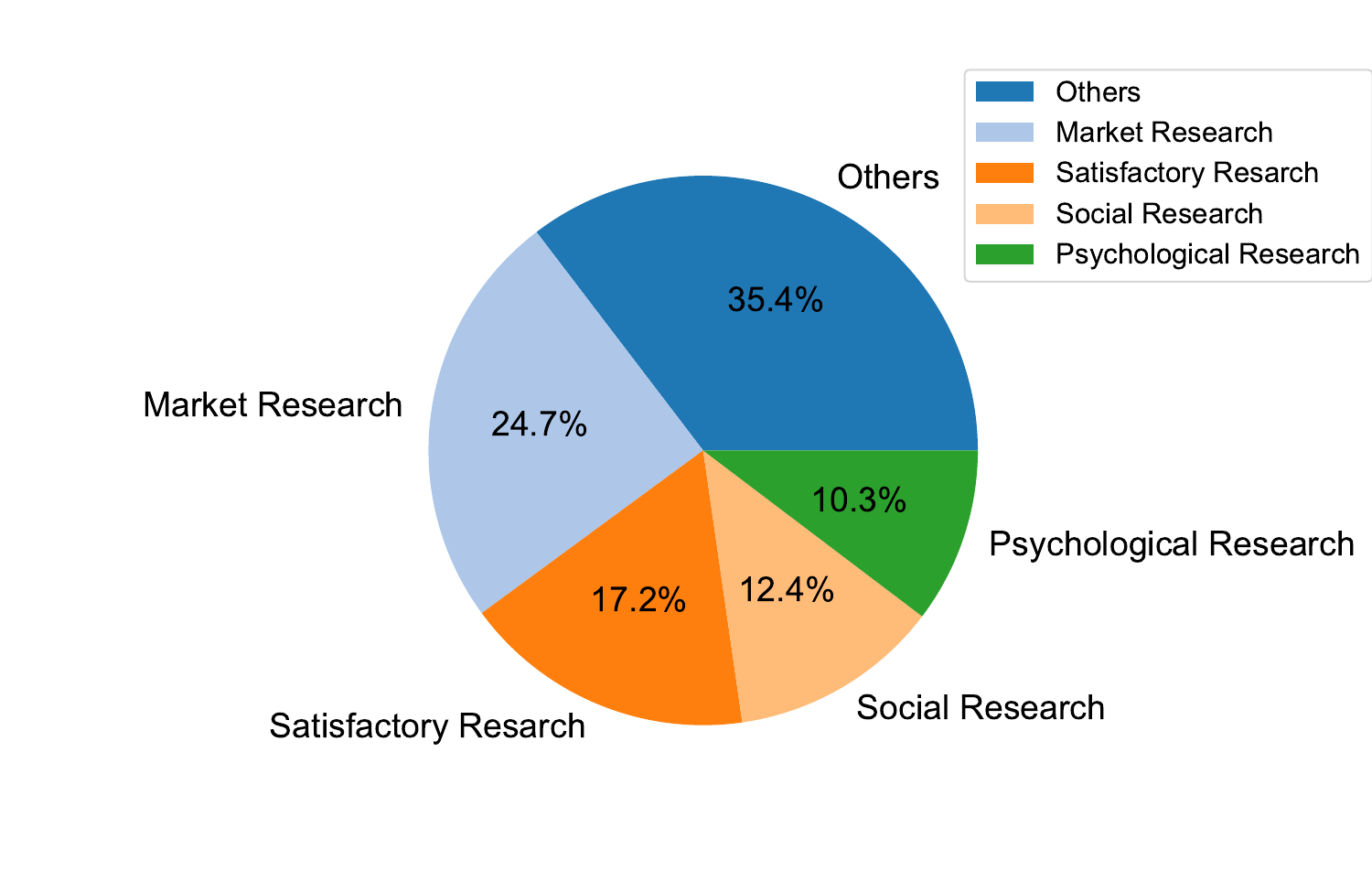}}
    \subcaption{\scriptsize{Type Distribution}}
    \label{fig: type category}
  \end{minipage}
  
  \caption{Visualization of Qsnail data statistics.}
  \label{fig:whole}
\end{figure}

\textbf{Question Analysis}. An examination of the Qsnail dataset reveals the following statistics: on average, there are approximately 15 questions for each questionnaire, as detailed in Table~\ref{tab: statistic}. To delve further into this, Figure~\ref{fig: question_num} illustrates the distribution of questionnaires based on the number of questions they contain. The dataset offers a substantial number of questions within each questionnaire, making it well-suited for the purpose of sequential question generation.

\textbf{Option Analysis}. In regard to individual questions within the questionnaires, the average number of options is more than four. Figure~\ref{fig: choice} provides a visual representation of the distribution of questions categorized by the number of available choices. Notably, the majority of questions offer \mbox{2 -- 7} options. The dataset provide adequate data to search for rational options generation.

\textbf{Domain Diversity}. The questionnaires in the dataset are grouped into 11 distinct application domains, encompassing areas like education, health, and internet. Figure~\ref{fig: topic category} provides an overview of the distribution of instances across these domains. Notably, each domain contains more than 100 samples, with education dominating the landscape, likely due to the prevalence of surveys among highly literate college students. Furthermore, Figure~\ref{fig: type category} outlines the distribution of instances across various research targets. The domains and survey purposes in the dataset are diverse, allowing for a comprehensive test of the effectiveness of questionnaire generation task.

\section{Experiments}
This section is dedicated to evaluating the performance of various models in questionnaire generation and answering four research questions.


\subsection{Research Questions}
In the context of the questionnaire generation task, we formulate four distinct research questions:

\textbf{RQ1}: How consistent are automatic and human evaluation metrics?

\textbf{RQ2}: How do traditional retrieval models and generative models perform on the questionnaire generation task?

\textbf{RQ3}: How do large language models (LLMs) perform on the questionnaire generation task?
 
\textbf{RQ4}: Can the outline-first prompt and fine-tuning approaches improve the performance?

\subsection{Baseline Models}
Here, we summarize all the models implemented for experiments:

\textbf{Retrieval and Generative Models.} We employ the well-known BM25 model~\cite{robertson2009probabilistic}, a sparse retrieval function, to retrieve pertinent questionnaires from the training dataset based on the research topic and intents. The top-1 result is selected as the final questionnaire.

Chinese-GPT2~\cite{radford2019language}, a decoder-only generative model, is utilized. We finetune the model and employ a comparative decoding approach to generate sequential question-option pairs, with the research topic and intents as input. The experiments are conducted using PyTorch on 8 NVIDIA V100 GPUs.

\textbf{Large Language Models.} We utilize ChatGPT~\cite{DBLP:journals/corr/abs-2303-08774}, a commercial LLM trained by reinforcement learning from human feedback, provided by the OpenAI API\footnote{https://openai.com/api/}. Additionally, we leverage Vicuna~\cite{vicuna2023}, an open-source LLM obtained by finetuning LLaMA on ShareGPT, with Vicuna-7B as the backbone model. Another baseline model, ChatGLM~\cite{DBLP:conf/iclr/ZengLDWL0YXZXTM23}, an open-source Chinese-English bilingual model, is employed with ChatGLM2-6B. We finetune language models (Vicuna-7B and ChatGLM-6B) with ZeRO-2 to distribute the model across 2 NVIDIA A100 (80G) GPUs. We set the learning rate, batch size, and maximum context length to $2 \times 10^{-5}$, 128, and 2048, respectively. All models are trained for 3 epochs.
\subsection{Evaluation Metrics}
In light of the constraints in questionnaire generation, we have developed comprehensive automatic and human evaluation metrics at the question, option, and overall levels. Given that model-generated contexts are often lengthy and not easily evaluated at a fine-grained level, manual extraction proves to be costly. In contrast, ChatGPT offers an excellent and cost-effective solution for information extraction~\cite{wei2023zero, jethani2023evaluating}. With the assistance of ChatGPT, we successfully split questions and options from the questionnaire. The final manual check confirms that the extracted content is consistent with the original content, satisfying the requirements.

\subsubsection{Automatic Evaluations}
Regarding the quality of questionnaire generation, we measured various representative indicators at the question level, option level, and overall level.

\textbf{Question-level.}
The quality of questions hinges on their relevance and specificity. To evaluate relevance, we measure the similarity between questions and the research topic and intent, both at the word and semantic levels. For word-level comparison, we use \mbox{Rouge-L}~\cite{lin2004rouge}, and for semantic-level comparison, we compute the cosine similarity. We utilize Sentence-Transformers~\cite{reimers2019sentence} for sentence-level embeddings, specifically the model \mbox{symanto/sn-xlm-roberta-base-snli-mnli-anli-xnli}\footnote{https://huggingface.co/symanto/sn-xlm-roberta-base-snli-mnli-anli-xnli}, trained for zero-shot and few-shot text classification. Given research topic $T$, research intents $I$, and sequential questions $q_i$, the similarity is calculated as follows:

$$\textrm{Cohen-sem} = \sum_{i=1}^m \textup{sim} \left(T \circ I, q_i\right). $$

In terms of specificity, we gauge it by examining word-level and semantic-level repetition between questions. The degree of repetition reflects the degree of specificity, and we calculate word-level repetition using the proportion of duplicate \ngrams in generated sequential questions:

$$
\textrm{Rep-n}= (1 - \frac{|\textrm{unique \ngrams}({q_1 \circ q_2 \cdots \circ q_m)}|}{|\textrm{total \ngrams}({q_1 \circ q_2 \cdots \circ q_m)}|}),
$$

where $q_1 \circ q_2 \ldots \circ q_m$ represents the sequence of all questions, $m$ is the total number of questions in the questionnaire. We group the input text into words based on the number of tokens, with each word consisting of $n$ tokens ($n \in \{2, 3, 4\}$). $\textrm{unique \ngrams}$ refers to words obtained after removing duplicates from all \ngrams.
Rep-n ranges from 0 (no repeating \ngrams) to 1.0 (maximum repetition).
In addition, higher values of the \textit{Diversity} \cite{su2022contrastive, li2022contrastive} indicate lower repetition.
At the semantic level, we introduce \textit{Rep-sem} to measure semantic repetition in questions. Similar to BertScore~\cite{zhang2019bertscore}, for each question, we select the most similar question and determine if it is a duplicate:

$$\mathrm{Rep}\mbox{-} \mathrm{sem} = \frac{\sum_{i=2}^m \mathbb{I}\left[\max _{j \le i-1} \textrm{Sim} \left(q_i, q_j\right) > \alpha \right]}{|m|},$$

$\mathbb{I}[·]$ is an indicator function that yields a value of 1 only when the similarity between two questions exceeds a certain threshold ($\alpha = 0.95$), and then the two questions are deemed to be duplicated.

\textbf{Option-level} The quality of options necessitates rationality, including relevance to the question, mutual exclusivity, completeness, and proper order. However, these aspects are challenging to assess through automated metrics. Similar to the relevance calculation for questions and research intents, we evaluate the relevance between options using Cohen-sem equation previously shown. The overlap between options and questions is almost nonexistent, so we do not consider word-level metrics.

\textbf{Overall-level} The sequence of questions should be well-ordered and align with the research intents. While the former cannot be measured through automated evaluation metrics, the latter is evaluated using BLEU-n~\cite{papineni2002bleu} to compute \ngrams matching scores between the generated text and human-written text. The greater the similarity between the generated questionnaire and the human-written text, the closer it aligns with the requirements.

\begin{table*}
\caption{Automatic evaluation results of models with different inputs on the Qsnail dataset.  $\uparrow$ means higher is better and $\downarrow$ means lower is better. `T' denotes the research topic, `I' denotes the research intents, `O' denotes the additional generated outline, and `Finetune' denotes models that are fine-tuning on datasets. Bold indicates the best results for the corresponding metric in all models except humans.}
\label{tab: auto}
\adjustbox{max width=\textwidth}{\begin{tabular}{llccccccccr}
\toprule
\multicolumn{1}{c}{\multirow{2}{*}{\textbf{Model}}} & \multicolumn{1}{c}{\multirow{2}{*}{\textbf{Method}}} & \multicolumn{5}{c}{\textbf{Question}} & \multicolumn{1}{c}{\textbf{Option}}                                                                       & \textbf{Overall}           \\ 
\cmidrule(lr){3-7} \cmidrule(lr){8-8} \cmidrule(lr){9-9}
\multicolumn{1}{c}{}                                & \multicolumn{1}{c}{}     & \textbf{Rouge-L $\uparrow$} & \textbf{Cohen-sem $\uparrow$ }                            & \textbf{Rep-2 / 3 / 4 $\downarrow$} & \textbf{Rep-sem $\downarrow$} & \textbf{Diversity $\uparrow$}   & \textbf{Cohen-sem $\uparrow$} & \textbf{BLEU-1 / 2 / 4 $\uparrow$}   \\
\midrule 
Human      &   -     & 8.54     & 34.54     & 26.69 / 12.92 / 7.28  & 8.62    & 59.19     & 25.75     & 100.0 / 100.0 / 100.0 \\
\hline
BM25       & T+I     & 5.33     & 20.81     & \textbf{32.58 / 19.25 / 13.78} & \textbf{7.58}    & \textbf{46.93}     & 20.89     & 31.19 / 19.89 / 10.24 \\
\hline
GPT-2      & Finetune     & 6.28        & 28.05         & 50.00 / 39.30 / 32.85            & 23.80       & 20.38         & 16.99         & 19.85 / 11.47 / 4.31 \\
\hline
\multirow{4}{*}{Vicuna-7B}  & T      & 8.74     & 39.71     & 74.41 / 68.76 / 64.20 & 38.99   & 2.86      & 20.86     & 22.08 / 12.39 / 4.57  \\
           & T+I     & 9.07     & 38.51     & 63.82 / 55.79 / 49.85 & 26.13   & 8.02      & 19.70     & 28.39 / 18.06 / 8.32  \\
           & T+I+O    & 9.39     & 38.30     & 63.87 / 55.40 / 49.12 & 20.81   & 8.19      & \textbf{33.82}     & 29.26 / 18.59 / 8.33  \\
           & Finetune     & 7.16        & 26.65         & 77.77 / 73.31  / 70.25             & 38.19       & 1.76         & 22.21         & 19.36 / 12.63 / 6.32  \\
\hline
\multirow{4}{*}{ChatGLM-6B} & T      & 7.20     & 38.39     & 46.53 / 35.29 / 28.73 & 13.14   & 24.65     & 26.20     & 31.90 / 18.33 / 6.62  \\
           & T+I     & 8.67     & 38.91     & 46.54 / 34.66 / 27.25 & 9.66    & 25.41     & 24.42     & 36.48 / 23.43 / 10.73 \\
           & T+I+O    & 8.45     & 37.13     & 47.27 / 35.96 / 28.51 & 11.33   & 24.14     & 23.75     & 33.23 / 20.68 / 9.01  \\
           & Finetune     & 7.29        & 32.34         & 45.53  / 33.15  / 26.09             & 10.37       & 26.91         & 18.85         & 34.61 / 22.64 / \textbf{11.09}  \\
\hline
\multirow{3}{*}{ChatGPT}    & T      & 8.84     & \textbf{46.13}     & 55.04 / 44.85 / 37.50 & 10.60   & 15.49     & 18.55     & 31.19 / 19.89 / 10.24 \\
           & T+I     & \textbf{11.99}    & 43.19     & 39.41 / 27.99 / 20.56 & 13.21   & 34.66     & 20.43     & 29.25 / 19.90 / 9.82  \\
           & T+I+O    & 9.33     & 40.50     & 42.31 / 30.17 / 22.55 & 7.88    & 31.20     & 22.80     & \textbf{36.74 / 23.91 /} 10.71    \\
\bottomrule
\end{tabular}}
\end{table*}

\begin{table*}
\caption{Human evaluation results of language models on Qsnail dataset. `T' denotes the research topic, `I' denotes the research intents, `O' denotes the additional generated outline, and `Finetune' denotes models that are fine-tuning on datasets. Bold indicates the best results for the corresponding metric in all models except humans.}
\centering
\label{tab: human}
\adjustbox{max width=0.85\textwidth}
{\begin{tabular}{llcccccc}
\toprule
\multicolumn{1}{c}{\textbf{Model}} & \multicolumn{1}{c}{\textbf{Method}} & \textbf{Relevance}$\uparrow$ & \textbf{Specifity}$\uparrow$ & \textbf{Rationality}$\uparrow$ & \textbf{Order}$\uparrow$ & \textbf{Background}$\uparrow$ & \textbf{Accept Rate}$\uparrow$  \\ 
\midrule
Human          & -               & 4.92                  & 4.94                 & 4.96                    & 4.94              & 4.56                   & 0.90                    \\
\hline
BM25          & T+I              & 1.88                  & \textbf{4.52}                 & \textbf{4.48}                    & \textbf{4.52}              & 4.04                   & 0.08                    \\
\hline
GPT-2          & Finetune              & 1.62                  & 1.68                 & 1.54                    & 1.52              & 1.84                   & 0.00                    \\
\hline
\multirow{3}{*}{Vicuna-7B}      & T               & 3.40                & 2.58               & 2.68                 & 2.68           & 1.06                & 0.08                 \\
               & T+I              & 4.24               & 3.38               & 3.36                 & 3.56           & 1.40                 & 0.20                  \\
               & T+I+O             & 4.30                & 3.62               & 2.66                 & 3.78           & 2.08                & 0.18                 \\
               & Finetune              & 1.98                   &  2.04                  & 2.00                     &  1.98              &  2.28                   &   0.02                   \\
\hline
\multirow{3}{*}{ChatGLM-6B}     & T               & 3.48               & 3.32               & 2.88                 & 3.08           & 2.52                & 0.10                  \\
               & T+I              & 4.32               & 3.78               & 3.32                 & 3.62           & 2.88                & 0.14                 \\
               & T+I+O             & 4.28               & 3.76               & 3.28                 & 3.68           & 3.34                & 0.20                  \\
               & Finetune              & 4.10                   & 3.86                   &  3.64                    &  3.60              &  3.50                   &  0.50                    \\
\hline
\multirow{3}{*}{ChatGPT}        & T               & 3.88              & 3.58              & 3.34                & 3.40          & 2.08               & 0.18                \\
               & T+I              & 4.14              & 3.46              & 3.42                & 3.48          & 2.36               & 0.20                \\
               & T+I+O             & \textbf{4.66}              & 4.12              & 3.80                & 3.82          & \textbf{4.18}     & \textbf{0.52} \\
\bottomrule         
\end{tabular}}
\end{table*}

\subsubsection{Human Evaluations}

In this study, we randomly sample 50 cases from the test set and engage three graduate annotators. Each annotator is presented with responses from various sources, including BM25, GPT-2, ChatGPT, ChatGLM-6B, Vicuna-7B, and a human source. These responses are intentionally shuffled to ensure anonymity. The annotators are asked to rate the questionnaires from the following six aspects:

\textbf{Relevance}. This dimension assesses the alignment of the questions with the research goals. Questions that strongly align with the research goals are deemed valuable, while those unrelated to the research goals should be excluded. The relevance indicator for a questionnaire is higher when a larger proportion of questions is pertinent to the research topic and intent. We map the ratio to the interval 1 to 5.

\textbf{Specificity}~\cite{lietz2010research,martin2006survey}. This aspect evaluates the extent to which questions in the questionnaire are specific. The primary function of a questionnaire is to transform broad research topics and intentions into precise, unambiguous questions. High specificity is achieved when a significant portion of the questions is detailed and specific. We map the ratio to the interval 1 to 5.

\textbf{Rationality}. This dimension examines the percentage of questions with logical and reasonable options. The options should align with the questions and adhere to constraints like mutual exclusive, complete, and logical. A higher proportion of questions with rational options results in a higher rationality indicator for the questionnaire. We map the ratio to the interval 1 to 5.

\textbf{Order}~\cite{taherdoost2022designing}. This dimension evaluates the logical flow and coherence of the question order in the questionnaire. Various guidelines are considered, such as transitioning from objective to subjective, from general to specific questions, or grouping similar questions together. A score of 1 indicates a very disorganized and distracting order, while 5 signifies a well-organized and logical sequence of questions.

\textbf{Background}. This dimension emphasizes the inclusion of background research questions, such as age and gender, which are crucial for maintaining the credibility of statistical data. Background questions serve as filters to exclude individuals who do not meet the study's requirements and can be used in data analysis. The adequacy of background questions contributes to a higher indicator. A score of 1 is assigned if there are almost no background questions, and 5 indicates the presence of comprehensive background questions.


\textbf{Acceptance Rate}. Finally, we introduce an overarching metric to represent the overall quality of the questionnaire. This metric assesses whether users are willing to adopt the questionnaire based on the research topic and intentions. A rating of 1 indicates a willingness to accept the questionnaire, while 0 signifies an unwillingness to accept it.

\subsection{Experimental Results}

We conduct experimental verification for the aforementioned raised questions.
\textbf{RQ1 - Consistency of automatic and human evaluations: } As depicted in Table~\ref{tab: auto}, at questions level, \mbox{Rouge-L} and \mbox{Cohen-sem} within automatic metrics align closely with `Relevance' in human evaluations. They consistently measure the degree of relevance of questions with respect to the research topic and intents. 
Similarly, \mbox{Rep-n}, \mbox{Rep-sem}, and diversity in automatic metrics, along with `Specifity' in human evaluations consistently reflect whether the questions are specific to the research topic and intents. 
However, when shifting to the options level, \mbox{Cohen-sem} within automatic metrics can only partially measure the similarity between options to the question, necessitating a supplementary evaluation of `rationality' assessment in the manual evaluation considering the intrinsic constraints, including mutually exclusive, complete, and orderly. 
At the overall level, the automatic evaluation metric \mbox{BLEU-n} coarsely measures the consistency between the generated questionnaire and the reference. To comprehensively evaluate the arrangement of sequential questions, additional aspects such as `order' and `background' in manual evaluations should be considered. To sum up, although automatic indicators may not fully encapsulate questionnaire quality, they still exhibit a strong correlation with human evaluations, thus serving as a valuable reference.

\textbf{RQ2 - Traditional models performance:} Retrieval models have limitations when handling new topics, while traditional generative models face challenges in producing coherent and usable questionnaires, even after finetuning. As depicted in Table~\ref{tab: auto}, BM25 excels in scoring highly on aspects related to form structure (e.g. Rep-n, Rep-sem, and Diversity), owing to its retrieval of human-crafted questionnaires characterized by high self-consistency. However, it receives significantly lower scores in terms of relevance to research topics and intents, as indicated by \mbox{Rouge-L} and \mbox{Cohen-sem}. 
Traditional generative model, like GPT-2, even after finetuning, gets notably low scores in terms of relevance, specificity, and diversity in both automatic and manual evaluation metrics (see Table~\ref{tab: auto} and Table~\ref{tab: human}), with obvious deficiencies in terms of readability and usability. The small parameter size and insufficient pre-training data make it difficult for GPT-2 to accomplish such a complex questionnaire generation task.

\textbf{RQ3 - Large language models performance:} 
LLMs excel in terms of relevance but still lag significantly behind humans when it comes to diversity, specificity, rationality, and order. Most LLMs, including the standard `T+I' version of Vicuna-7B, ChatGLM-6B, and ChatGPT, exhibit competitive performance with humans when it comes to relevance, as indicated by metrics such as \mbox{Rouge-L} and \mbox{Cohen-sem} in Table~\ref{tab: auto}, and the Relevance in Table~\ref{tab: human}. However, even the best LLMs reveal substantial disparities when compared to humans in other metrics, for example, question diversity 34.66 vs. 59.19 and rationality 3.80 vs. 4.96. Nevertheless, it is worth noting that LLMs consistently generate questionnaires with a structured format that remains relevant to users' needs, mainly attributable to their pretraining on an extensive and diverse corpus.

\textbf{RQ4 - Explore further improvements: } To address the above issues, we introduce two approaches to enhance performance. The first involves a chain-of-though prompt method known as ``outline-first'', mirroring the human writing process by firstly crafting an outline and then specific content. The second is to finetune models on Qsnail. 
The outline-first prompt significantly improves the performance of ChatGPT (e.g. \mbox{Rep-n}, \mbox{Rep-sem}, Diversity, Specificity, and Background) but not evident in \mbox{ChatGLM-6B} as well as \mbox{Vicuna-7B}, it's worth noting that outlines incorporate only the corresponding questions without associated options, which significantly decrease on `Rationality', particularly for \mbox{Vicuna-7B}. In addition, comparing `T', `T+I+O' and 'T+I' versions of models, we can see that the more information the input contains, the better the performance is. (see Table~\ref{tab: human}).
Finetuning models on Qsnail has proved to be beneficial for \mbox{ChatGLM-6B} on `Specificity' and `Background', thanks to the infusion of an extensive domain-specific knowledge. In contrast, \mbox{Vicuna-7B}, not specifically trained for the Chinese corpus, exhibits a heavily increase in repetitions after finetuning, leading to a deterioration in the quality.

\section{Related Work}
\textbf{Traditional Question Generation} has been applied in various significant scenarios, including question answering~\cite{zhu2021adaptive,}, machine reading comprehension, and automated conversations. Traditional question generation was tackled by rule-based methods~\cite{hussein2014automatic, labutov2015deep}, e.g., filling handcrafted templates under certain transformation rules. With the emergence of data-driven learning approaches, neural networks (NN) have gradually taken the mainstream. \citet{du2017learning} pioneer neural network-based approaches by adopting the Seq2seq architecture. Many ideas are proposed since then to make it more powerful. Furthermore, enhancing the Seq2seq model into more complicated structures using adversarial training, and reinforcement learning has also gained much attention~\cite{yao2018teaching, kumar2018putting}. In addition to unstructured question generation types, there are also structured question generation, include knowledge-based~\cite{song2016question, liang2023prompting, guo2022dsm, liang2023prompting} and table-based~\cite{chemmengath2021topic}.
There are also some works performing traditional question generation under certain constraints, e.g., controlling the topic~\cite{ding2023maclasa} and difficulty of questions~\cite{hu2018aspect, gao2018difficulty}.

\textbf{Sequential Question Generation} is challenging and is regarded as a conversational QG task. Existing sequential question generation models mainly focused on modeling complex context dependencies and frequently occurred coreference between questions. Therefore, sequential question generation is more challenging than standalone question generation. \citet{gao2019interconnected} achieve the best performance that generates sequential questions via coreference alignment and conversation flow modeling. \citet{chai2020learning} design an answer-aware attention mechanism and generate questions in a semi-autoregressive way to capture context dependencies. 

Unlike sequential question generation in dialogues, sequential questions in questionnaires not only contain coreference but also intrinsic constraints, including questions, options, and overall levels. The interconnected constraints make this task more challenging and raise curiosity about how different models perform on this task.
\section{Conclusion and Future Work}
We introduce Qsnail, the first dataset designed for questionnaire generation, a sequential question generation task, encompassing a wide array of intricate constraints related to questions, options, and overall structure. This comprehensive dataset comprises 13,000 questionnaires and approximately 184,000 question-option pairs. Extensive evaluations involving models like ChatGPT have consistently underscored the challenges inherent in generating questionnaires, highlighting the need for substantial future investigation. With the introduction of Qsnail, we aim to inspire further inquiry, fostering a renewed focus on the creation of professional questionnaires.
\maketitleabstract


\section{Ethical Considerations}
In our work, we use existing LLMs to generate questionnaires, so we have the same concerns as other text generation. For example, there is a risk of generating toxic or biased language. To assess and improve the performance of our questionnaire generation, this paper introduces a novel questionnaire dataset obtained through web crawling. This dataset may contain individuals' personal names, but we have implemented a filtering mechanism using a keywords filtering mechanism to exclude data containing such private information. This approach ensures the preservation of privacy.

\section{Acknowledgements}
We thank the anonymous reviewers for their insightful comments and suggestions. This work was supported by the National Key R\&D Program of China (2022YFB3103700, 2022YFB3103704), the National Natural Science Foundation of China (NSFC) under Grants No. 62276248, U21B2046, 62172393, U1836206, and U21B2046, and the Youth Innovation Promotion Association CAS under Grants No. 2023111, Zhongyuanyingcai program-funded to central plains science and technology innovation leading talent program (No.204200510002), and Major Public Welfare Project of Henan Province (No.201300311200).
Thanks to Hanxing Ding, Kangxi Wu, Junkai Zhou, and Zihao Wei for their kind help, as well as to Kaike Zhang and Xiaojie Sun for their valuable suggestions for revising the paper.

\section{References}
\bibliographystyle{lrec-coling2024-natbib}
\bibliography{lrec-coling2024-example}

\end{document}